\documentclass[journal,twocolumn,letterpaper]{IEEEJERM}

\ifCLASSINFOpdf
 
\else

\fi

\usepackage{times,amsmath,epsfig}
\renewcommand\IEEEkeywordsname{Index Terms}
\usepackage{fancyhdr}
\usepackage{amsmath}
\usepackage{amsfonts}
\usepackage{amssymb}
\usepackage[utf8]{inputenc}
\usepackage{array}
\usepackage{graphicx}
\usepackage{url}
\usepackage{subfigure}
\usepackage{bm}
\usepackage{breqn}
\usepackage{xcolor}
\usepackage{soul}
\usepackage{amssymb}
\usepackage{flushend}
\usepackage{graphicx}
\usepackage{multirow}
\usepackage{booktabs}

\IEEEoverridecommandlockouts

\begin{document}

\title{Improving Transformer Performance for French Clinical Notes Classification Using Mixture of Experts on a Limited Dataset}

\author{Thanh-Dung Le,~\IEEEmembership{Senior Member,~IEEE,}
        Philippe Jouvet M.D., and
        Rita Noumeir Ph.D.,~\IEEEmembership{Member,~IEEE}
 }

\markboth{IEEE }
{T.D. Le \MakeLowercase{\textit{et al.}}: }

\twocolumn[
\begin{@twocolumnfalse}
  
\maketitle

\begin{abstract}

Transformer-based models have shown outstanding results in natural language processing but face challenges in applications like classifying small-scale clinical texts, especially with constrained computational resources. This study presents a customized Mixture of Expert (MoE) Transformer models for classifying small-scale French clinical texts at CHU Sainte-Justine Hospital. The MoE-Transformer addresses the dual challenges of effective training with limited data and low-resource computation suitable for in-house hospital use. Despite the success of biomedical pre-trained models such as CamemBERT-bio, DrBERT, and AliBERT, their high computational demands make them impractical for many clinical settings. Our MoE-Transformer model not only outperforms DistillBERT, CamemBERT, FlauBERT, and Transformer models on the same dataset but also achieves impressive results: an accuracy of 87\%, precision of 87\%, recall of 85\%, and F1-score of 86\%. While the MoE-Transformer does not surpass the performance of biomedical pre-trained BERT models, it can be trained at least 190 times faster, offering a viable alternative for settings with limited data and computational resources. Although the MoE-Transformer addresses challenges of generalization gaps and sharp minima, demonstrating some limitations for efficient and accurate clinical text classification, this model still represents a significant advancement in the field. It is particularly valuable for classifying small French clinical narratives within the privacy and constraints of hospital-based computational resources.

\vspace{9pt}

\vspace{9pt}
\textbf{\textit{Clinical and Translational Impact Statement---} This study highlights the potential of customized MoE-Transformers in enhancing clinical text classification, particularly for small-scale datasets like French clinical narratives. The MoE-Transformer's ability to outperform several pre-trained BERT models marks a stride in applying NLP techniques to clinical data and integrating into a Clinical Decision Support System in a Pediatric Intensive Care Unit. The study underscores the importance of model selection and customization in achieving optimal performance for specific clinical applications, especially with limited data availability and within the constraints of hospital-based computational resources.}
\end{abstract}

\end{@twocolumnfalse}]

{
  \renewcommand{\thefootnote}{}%
  \footnotetext[1]{This work was supported in part by the Natural Sciences and Engineering Research Council (NSERC), in part by the Institut de Valorisation des donnees de l'Universite de Montreal (IVADO), in part by the Fonds de la recherche en sante du Quebec (FRQS), and in part by the Fonds de recherche du Quebec-Nature et technologies (FRQNT). }
  
  \footnotetext[2]{Thanh-Dung Le is with the Biomedical Information Processing Lab, Ecole de Technologie Superieure, University of Quebec, Canada, and also is with the Interdisciplinary Centre for Security, Reliability, and Trust (SnT), University of Luxembourg, Luxembourg (Email: thanh-dung.le@uni.lu).}

  \footnotetext[3]{Rita Noumeir is with the Biomedical Information Processing Lab, Ecole de Technologie Superieure, University of Quebec, Canada.}


  \footnotetext[3] { Philippe Jouvet is with the Research Center at CHU Sainte-Justine, University of Montreal, Canada.}
}
 
\IEEEpeerreviewmaketitle

\section{Introduction}
\IEEEPARstart{R}ecent advancements in deep learning have led to the development of Transformer models \cite{vaswani2017attention}, which have shown remarkable performance in various natural language processing (NLP) tasks \cite{tripathy2021comprehensive}. As a result, there is a growing interest in applying Transformer-based models to clinical applications, such as predicting disease risk \cite{huang2022assessing}, identifying disease \cite{ilias2022explainable}, and improving clinical decision-making \cite{meng2021bidirectional}. These models can be trained on various data sources, including electronic health records (EHRs) \cite{blanco2021exploiting}, and medical imaging \cite{deng2022rformer}, electroencephalogram \cite{phan2022sleepTransformer}, to extract relevant information and provide accurate predictions. Overall, Transformer models present a powerful tool for clinical applications and can potentially play an increasingly important role in healthcare.

In clinical NLP, Transformer models have shown great promise in clinical narrative classification. In this context, clinical narrative refers to patient encounters in EHRs or other clinical documentation. Using Transformer models, researchers and clinicians can develop algorithms that automatically classify these narratives based on different criteria, such as diagnosis, treatment, or patient outcomes. This can help streamline clinical workflows and improve patient care by providing more accurate and efficient clinical data processing. Some examples of successful applications of Transformer models for clinical narrative classification include identifying clinical coding \cite{lopez2021Transformers}, diagnosing health conditions \cite{ rizwan2022depression}, and detecting clinical events \cite{ kim2023identifying}. As such, Transformers-based models have become an increasingly important tool in clinical NLP and are likely to continue playing a significant role.

Despite their many benefits, Transformer models for clinical text classification have some limitations that must be considered. One major challenge is the need for large amounts of annotated clinical data to train these models effectively. Clinical data is often scarce and sensitive, which makes it challenging to obtain and annotate in a way that preserves patient privacy \cite{gao2021limitations}. Additionally, clinical language is highly specialized and can vary significantly across different specialties and regions, making it difficult to develop models that generalize well across different contexts \cite{bear2021clinically}. There is a risk of bias in the data used to train these models, leading to errors or disparities in the predictions made \cite{alimova2021cross}. Furthermore, the computational requirements of Transformer-based models can be pretty high, which can limit their use in resource-constrained settings where computational resources are limited \cite{gillioz2020overview}. Finally, the interpretability of these models can be limited, making it difficult for clinicians to understand how they make their predictions and trust their outputs \cite{rudin2019stop}. While Transformer models have great potential for clinical text classification, they also require careful attention to their limitations and the potential biases that can arise.

Additionally, engaging with real clinical data presents various challenges and constraints. A primary issue is the limited availability of clinical data, compounded by its inherently confidential nature. In addition, computational resources are constrained as we rely on a shared cloud computing infrastructure hosted on the hospital server. The hospital's server is the sole environment permitted for data processing and suffers from limited computational capacity. This presents a significant bottleneck, especially when developing sophisticated machine learning algorithms that demand high computational power to provide real-time analysis and predictive insights. Another challenge is that clinical datasets are often small and imbalanced, making it difficult to train accurate models using Transformer \cite{neveol2018clinical}. Small datasets can also lead to overfitting, where the model performs well on the training data but fails to generalize to new data. When there is insufficient data, the Transformer model does not learn to focus on local features in the lower layers of the network. This may result in reduced model performance, as it cannot effectively capture relevant information from the input data \cite{raghu2021vision}. Overall, while Transformer models offer many advantages for clinical text classification, their effectiveness is influenced by the data's language and the training dataset's size and quality.

This study aims to overcome the abovementioned challenges using Transformer-based models for clinical text classification for a small French clinical note and constrained computation resources. Motivated by the recent Switch Transformer model developed by Google \cite{fedus2021switch}, we will adapt the Mixture of Experts (MoE) mechanism to improve the performance of the Transformer model, namely the MoE-Transformer. A MoE-Transformer is an extension of the Transformer architecture motivated by the original model's self-attention mechanisms. Still, it uses an MoE mechanism to address the limitations of the conventional Transformer \cite{vaswani2017attention}. A key technical difference between MoE-Transformers with an MoE mechanism and Transformers with self-attention is how they model complex input-output relationships. An example of the effectiveness of MoE has been proven by \cite{xue2022go}; that study shows that the approach of using parameter sharing to compress along the depth of the model, which is used in existing works, is limited in terms of performance. To improve the model's capacity, the authors propose scaling along the model's width by replacing the feed-forward network with an MoE layer. This allows for better modeling capacity and potentially better performance. 

Additionally, the study \cite{lazaridou2021mind} suggests that simply increasing the model's size is insufficient to address the issue of performance degradation over time from neural language models. However, the researchers found that using models that continuously update their knowledge with new information can help alleviate this problem. While Transformers with self-attention model these relationships through a single attention mechanism that captures dependencies between all input and output positions, MoE-Transformers with an MoE mechanism decompose the problem into smaller, simpler sub-problems, each handled by a different ``expert" model. In other words, instead of using a single global attention mechanism, MoE-Transformers employ multiple local attention mechanisms focusing on different input aspects. Depending on the context, the gating mechanism used in MoE-Transformers selects which expert model to use for a given input. Therefore, this approach can potentially improve the modeling of complex input-output relationships and increase the model's efficiency, especially when dealing with complex data from the clinical domain. This is particularly important in clinical data, where information is often conveyed through complex and nuanced language. By employing this approach, our study aims to improve the accuracy and generalizability of clinical text classification models for small datasets in languages other than English. We have made several significant contributions to clinical text classification using Transformer-based models. 
\begin{itemize}
    \item First, our study demonstrates a comprehensive implementation of a simplified MoE-Transformer model, and trained from scratch. This would allow other researchers to understand and replicate the methodology used in the study, which is essential for advancing this work.
    \item Second, our study provides experimental evidence showing the limitations of Transformer-based models regarding generalization gap and sharp minima. This highlights the importance of carefully training these models to avoid overfitting and improve generalization performance. 
    \item Finally, our study illustrates the interpretable output of the model by adapting the Integrated Gradients (IG) \cite{sundararajan2017axiomatic}. It provides a way to attribute importance to the input features of a model, allowing clinicians and researchers to gain insight into how the model is making its predictions.
    
\end{itemize}

The main contribution of this work is the introduction of a carefully curated, small-scale French clinical text dataset from CHU Sainte-Justine (CHUSJ) Hospital, which can serve as a valuable benchmark for resource-constrained NLP research in healthcare. Moreover, our findings demonstrate the feasibility of deploying the proposed MoE-Transformer in a practical clinical decision support system (CDSS), even when both the available data and computational resources are minimal.

\section{MATERIALS AND METHODS}
\label{sec:autoencoder}

\subsection{French Clinical Data at CHUSJ}

The CDSS system in the CHUSJ hospital aims to improve the diagnosis and management of acute respiratory distress syndromes (ARDS) in real-time by automatically screening data from electronic medical records, chest X-rays, and other sources. Previous studies have found that the diagnosis of ARDS is often delayed or missed in many patients \cite{bellani2016epidemiology}, emphasizing the need for more effective diagnostic tools. To diagnose ARDS, three main conditions must be detected: hypoxemia, chest X-ray infiltrates, and absence of cardiac failure \cite{pediatric2015pediatric}. The research team at CHUSJ has developed algorithms for detecting hypoxemia \cite{sauthier2021estimated}, analyzing chest X-rays \cite{zaglam2014computer, yahyatabar2020dense}, and identifying the absence of cardiac failure. In addition, the team has performed extensive analyses of machine learning algorithms for detecting cardiac failure from clinical narratives using natural language processing \cite{le2021detecting, le2023adaptation}. Implementing these algorithms could increase ARDS diagnosis rates and improve patient outcomes.

This study was conducted following ethical approval from the research ethics board at CHUSJ (protocol number: 2020-2253), and the study's design focused on identifying cardiac failure in patients within the first 24 hours of admission by analyzing admission and evolution notes during this initial period. The dataset consisted of 580,000 unigrams extracted from 5,444 single lines of short clinical narratives. Of these, 1,941 cases were positive (36\% of the total), and 3,503 cases were negative. While the longest n-gram was over 400 words, most n-grams had a length distribution between 50 and 125 words. The average length of the number of characters was 601 and 704, and the average size of the number of digits was 25 and 26 for the positive and negative cases, respectively. We pre-processed the data by removing stop-words and accounting for negation in medical expressions. Numeric values for vital signs (heart rate, blood pressure, etc.) were also included and decoded to account for nearly 4\% of the notes containing these values. All the notes are short narratives; detailed characteristics for the notes at CHUSJ can be found in the Supplementary Materials from the study \cite{le2021detecting}.

\begin{figure}[t]
	\centering
	\vspace{2pt}
	\includegraphics[scale=0.6]{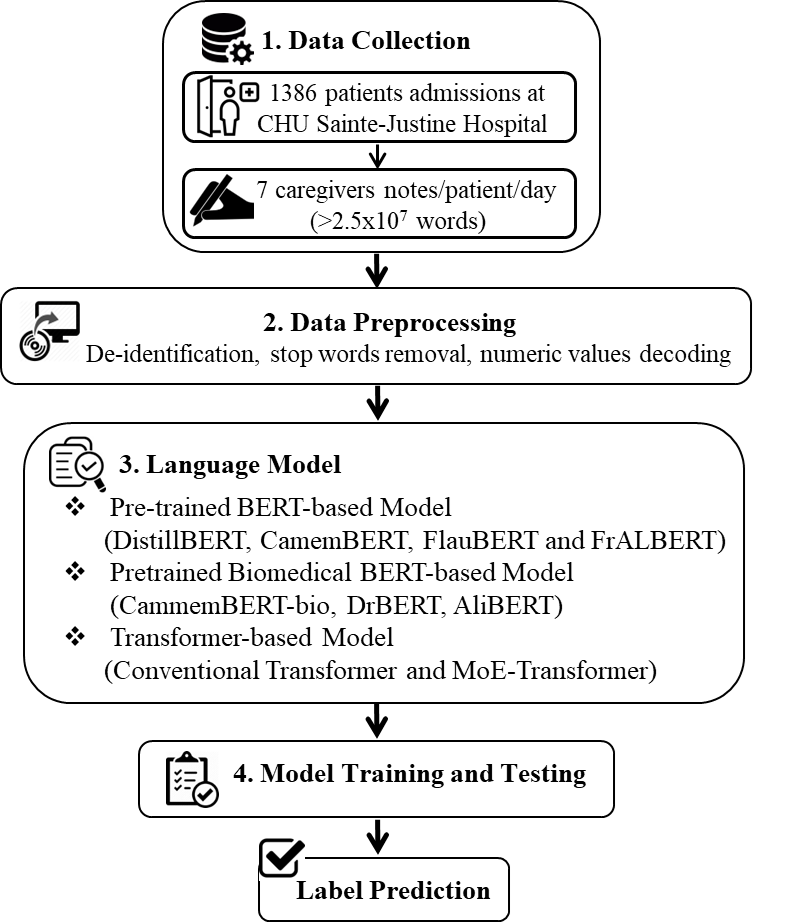}
	\vspace{-2pt}
	\caption{Workflow demonstration of the proposed methodology to classify French clinical narratives at CHUSJ hospital.}
	\vspace{-8pt}
	\label{fig:CHUSJ_workflow}
\end{figure}

\subsection{Language Models for Clinical Narratives}

This manuscript thoroughly analyzes the present state of pre-trained BERT-based models and Transformer models for clinical narrative classification, with a particular emphasis on limited datasets. Various pre-trained BERT-based models for the French language are leveraged, such as FlauBERT, FrALBERT, CamemBERT, and DistilBERT, as depicted in Fig. \ref{fig:CHUSJ_workflow}. Moreover, conventional and MoE-Transformer models are constructed from scratch to perform the same task. Finally, we compare the performance of all models based on various evaluation metrics for binary classification, including accuracy, precision, recall, F1-score, and area under the curve (AUC). This study endeavors to offer insights into the efficacy of these models on limited datasets, which is a critical aspect in real-world clinical settings for non-English notes.

\begin{figure*}[!ht]
	\centering
	\vspace{2pt}
	\includegraphics[scale=0.45]{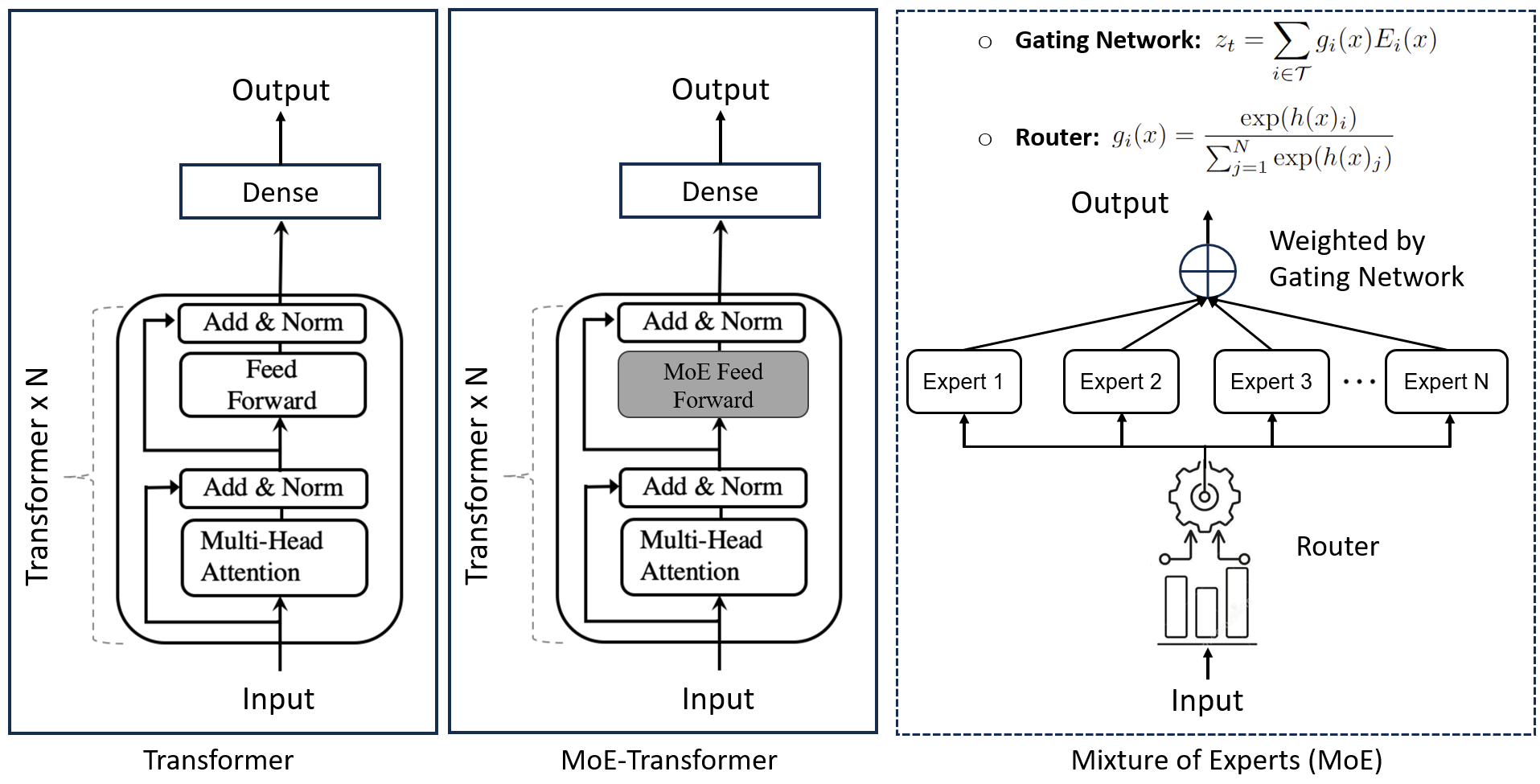}
	\vspace{-2pt}
	\caption{Illustration of a Conventional Transformer \cite{alammar2018illustrated} (left), an MoE-Transformer (middle), and the detailed MoE block (right).}
	\vspace{-8pt}
	\label{fig:trans_vs_switch}
\end{figure*}

\subsubsection{Transformer-based Models}
Transformer-based models have been highly effective for various NLP tasks, including text classification. The conventional Transformer model \cite{vaswani2017attention} with multi-head self-attention is a widely used architecture for this task. Shown in Fig. \ref{fig:trans_vs_switch} (left), its architecture comprises an encoder consisting of multiple layers of multi-head self-attention and feedforward neural networks (FFN). The multi-head self-attention mechanism allows the model to weigh the importance of different words in a sequence based on their semantic relationships, while the FFNs transform the output of the self-attention layer into a more helpful representation. The Transformer's core is the self-attention mechanism based on mathematical expressions \cite{lin2022survey}. Given a sequence of input embeddings $x_1, ..., x_n,$ the self-attention mechanism computes a set of context-aware embeddings $h_1, ..., h_n$ as follows:

\begin{align}
    {h}_i = \text{Attention}(QW_i^Q,KW_i^K,VW_i^V)
    \label{eq:head_i}
\end{align}
    
where $\text{Attention}$ is the scaled dot-product attention function:
\begin{align}
    \text{Attention}(Q,K,V) = \text{softmax}\left(\frac{QK^T}{\sqrt{d_k}}\right)V
    \label{eq: Attetion}
\end{align}
Then, the multi-head attention is a concatenation of all head of $h_i$, as follows:
\begin{align}
    \text{MultiHead}(Q,K,V) = \text{Concat}(h_1,\ldots,h_n)W^O
    \label{eq: concat}
\end{align}

Additionally, the position-wise FFNs are multi-layer perceptrons applied independently to each position in the sequence, which provide a nonlinear transformation of the attention outputs. FNNs are calculated as follows:
\begin{align}
    \text{FFN}(x) = \text{ReLU}(xW_1+b_1)W_2+b_2
    \label{eq: ffn}
\end{align}

For each layer, there is a Layer Normalization which normalizes the inputs to a layer in a neural network to improve training speed and stability.
\begin{align}
    \text{LayerNorm}(x) = \gamma\frac{x-\mu}{\sqrt{\sigma^2 + \epsilon}} + \beta
\end{align}

\noindent where $Q$, $K$, and $V$ are the query, key, and value matrices, $W_i^Q$, $W_i^K$, and $W_i^V$ are the learned weight matrices for the $i$-th head of the multi-head attention, $W_1$ and $W_2$ are the weight matrices for the position-wise FFNs, $\gamma$ and $\beta$ are learned scaling and shifting parameters for layer normalization, and $\mu$ and $\sigma$ are the mean and standard deviation of the input feature activations.

By performing these steps for each layer in the encoder and decoder, the multi-head self-attention mechanism allows the Transformer architecture to capture rich semantic relationships between different words in a sequence and is highly effective for a wide range of NLP tasks. However, the conventional Transformer architecture has some limitations. One of the main issues is that the self-attention mechanism requires quadratic computation time concerning the input sequence length, making it difficult to scale the model to very long sequences \cite{raffel2020exploring}, and lower generalizability for a short sequence \cite{gao2021limitations}. Additionally, the self-attention mechanism treats all positions in the input sequence equally, which may not be optimal for certain types of inputs where some positions are more critical than others. While the Transformer model has shown great performance, it can still struggle to capture complex input-output relationships requiring more specialized models.

Motivated by study \cite{fedus2021switch}, the MoE structure attempts to address these limitations. The MoE mechanisms decompose the problem into smaller, simpler sub-problems, allowing the model to handle sequences and complex input-output relationships better. As mentioned earlier, the multi-head self-attention mechanism in the Transformer model is motivated by the need to capture semantic relationships between words in a sequence, but it has limitations when dealing with short sequences \cite{gao2021limitations}. The MoE mechanisms allow the model to divide the sequence into smaller, more manageable segments and apply different experts to each segment \cite{dikkala2023benefits}. Hence, it can overcome the limitations by increasing the capacity of the FFNs of conventional Transformers \cite{chen2022towards}. Consequently, this approach has improved the model's sequence task classification performance and achieved state-of-the-art results on several benchmarks \cite{xue2022go, lazaridou2021mind, fan2021beyond}. 

The critical difference in the mathematical equation of the MoE-Transformer compared to the conventional Transformer is replacing the FFN with the MoE mechanism, shown in Fig. \ref{fig:trans_vs_switch}. In the conventional Transformer (left), the FFN consists of two linear layers with a ReLU activation function in between. On the other hand, the MoE mechanism (right) uses a set of expert networks to learn different aspects of the input data and then combines their outputs with a gating network. It allows the model to dynamically choose between multiple sets of parameters (i.e., expert modules) based on the input. This contrasts the original Transformer model in Eq. \ref{eq: ffn}, which uses a fixed set of parameters for all inputs. Formally, the MoE mechanism in the MoE-Transformer can be represented by the following equation:

\begin{align}
    z_t = \sum_{i \in \mathcal{T}} g_i(x) E_i(x)
\end{align}

\noindent where $g_i(x_t)$ is a gating function that determines the importance of expert module i for input $x_t$, and $e_i(x_t)$ is the output of expert module i for input $x_t$. The top-$k$ gate values are selected for routing the token $x$. If $\mathcal{T}$ is the set of selected top-$k$ indices, then the output computation of the layer is the linearly weighted combination of each expert's computation on the token by the gate value.

Next the MoE layer will take as an input a token representation \( x \) and then routes this to the best-determined top-\( k \) experts, selected from a set \(\{ E_i(x) \}_{i=1}^{N}\) of \( N \) experts. The router variable \( W_r \) produces logits \( h(x) = W_r \cdot x \) which are normalized via a softmax distribution over the available \( N \) experts at that layer. The gate-value for expert \( i \) is given by,

\begin{align}
     g_i(x) = \frac{\exp(h(x)_i)}{\sum_{j=1}^{N} \exp(h(x)_j)}
\end{align}

The gating network mechanism is implemented by learning the parameters of the gating functions, which are used to select the expert modules dynamically. This allows the model to adapt to different input distributions and perform better on various tasks. Here is a summary of how the MoE mechanism works in the MoE-Transformer:

\begin{enumerate}
    \item  The input is split into multiple subspaces, and each subspace is processed by a separate expert. Each expert is a separate neural network trained to specialize in a specific subset of the input space.

    \item The output of each expert is a vector that represents its prediction for the given input subspace.

    \item A gating mechanism selects the most relevant expert for a given input. This gating mechanism takes the input and produces a set of weights that determine the importance of each expert's prediction.

    \item The final output is a weighted combination of the experts' predictions. The weights used in the combination are determined by the gating mechanism.

\end{enumerate}

Overall, the MoE allows the MoE-Transformer to learn complex patterns in the input space by leveraging the specialized knowledge of multiple experts. The MoE framework enables the model to learn from multiple experts, each specialized in different aspects of the data, and combine their outputs to achieve better performance. This can lead to better performance on tasks requiring understanding inputs and offers a promising solution to the challenge of small datasets in clinical text classification. Consequently, this study uses its ability to capture the complex relationships between words and phrases in the clinical text.

\subsubsection{Pre-trained BERT-based Models for French}
Pre-trained BERT-based models have become increasingly popular, enabling researchers and practitioners to perform various language-processing tasks with unprecedented accuracy. While BERT \cite{devlin2019bert} was initially developed for English language processing, it has since been adapted to several other languages, including French. In this context, we will explore some of the most popular pre-trained BERT-based models for French language processing available from Huggingface, including CamemBERT \cite{martin2020camembert}, FlauBERT \cite{le2020flaubert}, FrALBERT \cite{CattanSR21}, and DistillBERT \cite{sanh2019distilbert}.

\subsubsection{Pre-trained Biomedical BERT-based Models for French}
The biomedical pre-trained BERT-based models, including CamemBERT-bio \cite{corr2306}, DrBERT \cite{LabrakBDRMDG23}, and AliBERT \cite{BerheDMMDZ23}, are specifically designed for processing and understanding biomedical text. CamemBERT-bio is tailored for French biomedical data, leveraging the strengths of the CamemBERT architecture to provide robust performance in this domain. DrBERT and AliBERT also contribute to the landscape of specialized models, offering high accuracy and efficiency in various biomedical NLP tasks. These models are particularly well-suited for French clinical notes classification, as they have been trained on extensive French biomedical corpora, ensuring they capture the nuances and specific terminology used in French medical practice.

\section{Experimental Implementation}
\label{sec:exp_result}

Table \ref{tab:models-hyperpara} shows the total parameters of different Transformer-based models used in this study, including CamemBERT, DistillBERT, FlauBERT, FrALBERT, Transformer, and MoE-Transformer. The parameters compared include hidden layers, attention heads and total parameters. Regarding total parameters, CamemBERT, CamemBERT-bio, DrBERT and AliBERT have the highest number of parameters, with more than 110 million. While Transformer, and MoE-Transformer have significantly fewer parameters, with 2.3 million, and 5.7 million, respectively. The variation in parameters across different models reflects the differences in the architecture and design of the models. This information is crucial for understanding each model's computational complexity and efficiency and helps select the most suitable model.

\begin{table*}[t]
\footnotesize
\centering
\caption{Total parameters of the fine-tuned models}
\label{tab:models-hyperpara}
\begin{tabular}{|l|l|l|l|}
\hline
\multicolumn{1}{|c|}{Models} & Hidden Layers & Attention Heads & Total Parameters (Millions) \\ \hline
DistillBERT    & 6  & 12 & 67.2  \\ \hline
CamemBERT      & 12 & 12 & 111.4 \\ \hline
FlauBERT       & 6  & 8  & 54.9  \\ \hline
FrALBERT       & 12 & 12 & 12.7  \\ \hline
CamemBERT-bio  & 12 & 12 & 111.4 \\ \hline
DrBERT         & 12 & 12 & 111.4 \\ \hline
AliBERT        & 12 & 12 & 117.6 \\ \hline
Tranformer     & 4  & 4  & 2.9   \\ \hline
MoE-Transformer & 4  & 4  & 5.7   \\ \hline
\end{tabular}
\end{table*}

\begin{figure*}[t]
	\centering
	\vspace{2pt}
	\includegraphics[scale=0.5]{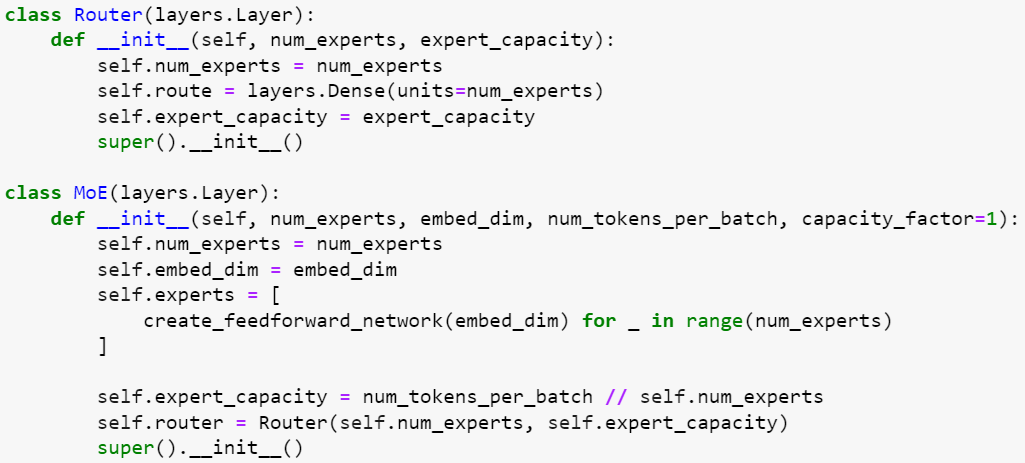}
	\vspace{-2pt}
	\caption{Pseudocode snippet in Python for the implementation of MoE layer.}
	\vspace{-8pt}
	\label{fig:MoE_Python}
\end{figure*}

Defining the hyperparameters during the training process of Transformers is a critical step in achieving good performance. Here are some of the critical hyperparameters that are tuned during the training process of BERT-based and Transformer models in this study:

\begin{itemize}
    \item \textbf{Maximum sequence length}: This is the maximum number of tokens that can be inputted into the model simultaneously. Setting an appropriate maximum sequence length can affect the performance and memory usage of the model. Due to computational constraints, the maximum sequence length varies from 128 to 256.

    \item \textbf{Batch size}: Choosing an appropriate batch size can affect the speed and stability of the training process. We varied the training batch size for each trial, ranging from 4 to 32 (with gradient accumulation as 4), based on the knowledge that training with smaller batches is more effective for highly low-resource language training \cite{atrio2021small}.
    
    \item \textbf{Drop-out}: This regularization technique randomly drops out some of the neurons during training to prevent overfitting. The dropout rate determines the proportion of neurons that drop out during each iteration \cite{srivastava2014dropout}.

   \item \textbf{Optimizers}: These algorithms update the model weights during training to minimize the loss function. Different optimizers have different strengths and weaknesses, and choosing the right one can impact the final performance of the model. Adaptive Moment Estimation (Adam) \cite{KingmaB14}, AdamW (Adam with weight decay) \cite{LoshchilovH19} were used.

    \item \textbf{Learning rate}: Consine annealed learning rate with warmup can help prevent training instability in the deeper layers of a neural network; its primary purpose is to help the model converge more quickly and effectively to a better solution overall \cite{otmareKXS19}.
     
    \item \textbf{Number of multi-head attention}: This determines the number of attention heads used in the multi-head attention layer of the Transformer. Increasing the number of attention heads can improve the model's ability to attend to different input parts.

    \item \textbf{Number of experts}: This determines the number of experts used in the MoE layer of the Transformer. Increasing the number of experts can improve the model's ability to handle diverse inputs. In implementing the MoE, we followed the guidelines from \cite{fan2024towards, zhao2024generalization}, and an example pseudocode snippet in Python for implementing the MoE layer shown in Fig. \ref{fig:MoE_Python}.
    
\end{itemize}

\begin{table*}[t]
\footnotesize
\centering
\caption{Hyperparameters of the fine-tuned models}
\label{tab:models-hyperpara-finetune}
\begin{tabular}{|l|l|l|l|}
\hline
Hyperparameters                & Pretrained BERT-based & Transformer  & MoE-Transformer \\ \hline
Number of multi-head attention & N/A                     & 4            & 4                  \\ \hline
Number of Experts              & N/A                     & N/A            & 4                  \\ \hline
Batch size                     & 16                    & 16           & 16                 \\ \hline
Dropout                        & 0.5                   & 0.35         & 0.35               \\ \hline
Learning rate                  & Cosine annealed          & Cosine annealed & Cosine annealed      \\ \hline
Optimizer                      & Adam                  & AdamW        & AdamW              \\ \hline
Adam\_$\epsilon$                  & N/A                     & 5*1e-06      & 5*1e-06            \\ \hline
Maximum sequence length        & 256                   & 256          & 256                \\ \hline
\end{tabular}
\end{table*}

Choosing appropriate values for these hyperparameters requires careful experimentation and tuning to achieve the best possible results. Only a few parameters require careful tuning. As reported in \cite{popel2018training}, the model size, learning rate, batch size, and maximum sequence length are the critical hyperparameters for Transformer model training. For this reason, grid search can be an efficient approach for optimizing these parameters by simultaneously exploring all possible combinations of intervals. The combination with the highest estimated performance was considered the optimal solution, and this approach balances computational efficiency and models' accuracy.

Finally, table \ref{tab:models-hyperpara-finetune} presents the hyperparameters used to fine-tune three models. For the pre-trained BERT-based model, the number of multi-head attention and the number of experts are not applicable (N/A), as this model is already trained and does not require further customization. The batch size, epochs, dropout rate, learning rate, and optimizer for all models are specified. The trained BERT-based model uses an Adam optimizer with a dropout rate 0.5 and a cosine decay learning rate. The Transformer and MoE-Transformer models use an AdamW optimizer with a dropout rate of 0.35 and a cosine decay learning rate. The Adam\_$\epsilon$ is only specified for the Transformer and MoE-Transformer models and is set to 5*1e-06. The maximum sequence length for all models is set to 256. Additionally, the GlorotNormal initializer \cite{glorot2010understanding}, batch normalization \cite{ioffe2015batch} are employed for models' stability. Then, these hyperparameters were carefully chosen to achieve optimal performance and prevent overfitting. 

The data was divided into 80\% training and 10\% validation and 10\% testing. To assess the performance of our method, metrics including accuracy, precision, recall, and F1 score were used \cite{goutte2005probabilistic}. These metrics are defined as follows. 
\begin{align}
&\text {Accuracy}=\frac{\mathrm{TP}+\mathrm{TN}}{\mathrm{TP}+\mathrm{TN}+\mathrm{FP}+\mathrm{FN}} \nonumber \\ 
&\text {Precision}=\frac{\mathrm{TP}}{\mathrm{TP}+\mathrm{FP}} \nonumber \\
&\text {Recall/Sensitivity}=\frac{\mathrm{TP}}{\mathrm{TP}+\mathrm{FN}} \nonumber \\ 
&\text {F1-Score} =\frac{2^{\star} \text {Precision}^{\star} \text {Recall}}{\text {Precision }+\text {Recall}} \nonumber
\end{align}

\noindent where TN and TP stand for true negative and true positive, respectively, and are the number of negative and positive patients correctly classified. FP and FN represent false positives and false negatives, and the number of incorrectly predicted positive and negative patients.

\section{Results and Discussion}
\label{sec:dis}
\begin{figure*}[t]
	\centering
	\vspace{2pt}
	\includegraphics[scale=0.5]{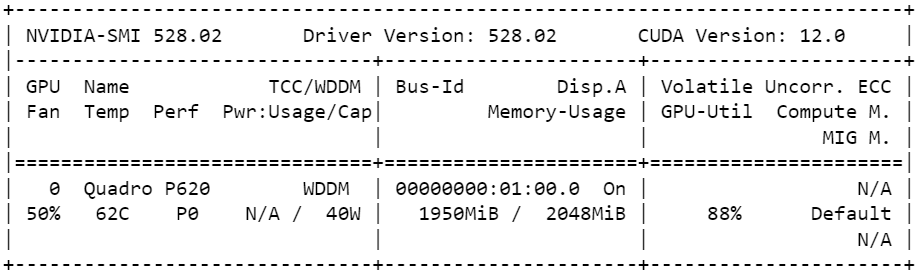}
	\vspace{-2pt}
	\caption{GPU utilization and resource usage during model training on NVIDIA Quadro P620.}
	\vspace{-8pt}
	\label{fig:cuda}
\end{figure*}

\begin{figure*}[t]
	\centering
	\vspace{2pt}
	\includegraphics[scale=0.6]{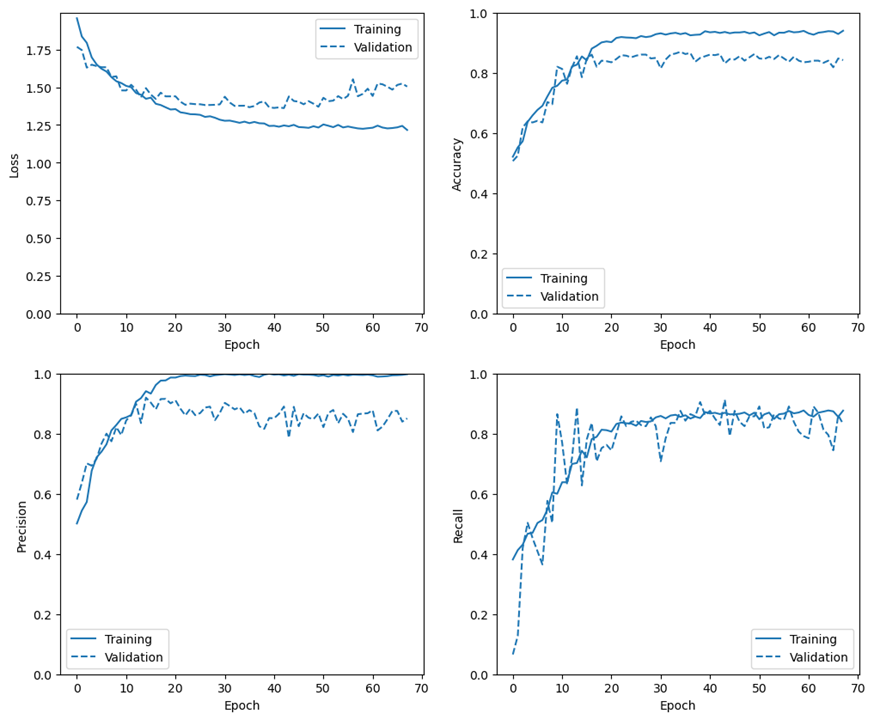}
	\vspace{-2pt}
	\caption{Training and validation performance results from MoE-Transformer model.}
	\vspace{-8pt}
	\label{fig:switch_results}
\end{figure*}

First, all models were trained using an NVIDIA Quadro P620 GPU. As shown in Fig. \ref{fig:cuda}, this GPU has a total memory capacity of 2048 MB, of which 1950 MB was actively utilized during training. The GPU utilization rate reached 88\%, indicating a high level of resource engagement. The Quadro P620, while robust, operates within relatively modest resource constraints compared to more powerful GPUs typically used for deep learning tasks. Training pre-trained models under such limited computational resources poses significant challenges. Efficient resource management and optimization techniques are crucial to achieve competitive performance without overloading the system.

During training and validation shown in Fig. \ref{fig:switch_results}, the MoE-Transformer model showed a gradual decrease in loss with increasing epochs. The loss started to converge after around 20 epochs and reached its minimum at the 30th epoch. Applying the early stopping at this point helped prevent the model's overfitting. The accuracy and precision of the model showed a smooth convergence to their optimal values for both the training and validation phases. However, the recall values for the two phases fluctuated quite a bit. The model's overall performance was good, with high accuracy, precision, and recall. The model's ability to reach its optimal values with smooth convergence and with the help of early stopping indicates the model's effectiveness in the given task.

\begin{figure}[t]
	\centering
	\vspace{2pt}
	\includegraphics[scale=0.576]{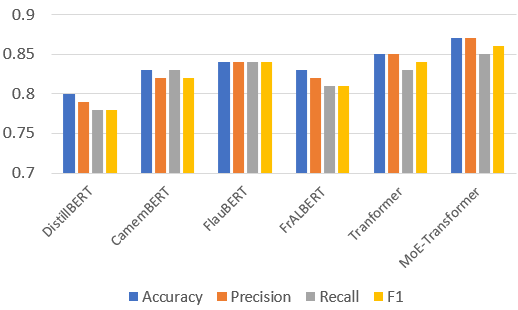}
	\vspace{-2pt}
	\caption{Performance comparison between MoE-Transformer vs. pre-trained BERT-based models, including DistillBERT, CamemBERT, FlauBERT, and FrALBERT.}
	\vspace{-8pt}
	\label{fig:MoE_Bert}
\end{figure}

\begin{figure}[t]
	\centering
	\vspace{2pt}
	\includegraphics[scale=0.625]{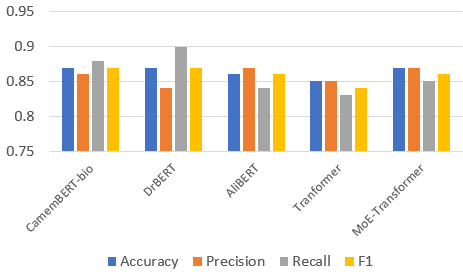}
	\vspace{-2pt}
	\caption{Performance comparison between MoE-Transformer vs. pre-trained biomedical BERT-based models, including CamemBERT-bio, Dr.BERT, and AliBERT.}
	\vspace{-8pt}
	\label{fig:MoE_BioBert}
\end{figure}

Based on the performance comparison presented in Table \ref{tab:model_comparisions}, the MoE-Transformer demonstrates a significant advantage in computational efficiency while maintaining competitive performance metrics compared to pre-trained BERT-based and biomedical BERT-based models. Specifically, in Fig. \ref{fig:MoE_Bert}, the MoE-Transformer achieves high scores across all performance metrics: accuracy (0.87), precision (0.87), recall (0.85), and F1 score (0.86). These results are comparable to or better than those of pre-trained BERT-based models such as DistillBERT, CamemBERT, FlauBERT, and FrALBERT. Notably, the MoE-Transformer outperforms CamemBERT and FrALBERT in all metrics. Fig. \ref{fig:MoE_BioBert} demonstrates that the MoE-Transformer performs competitively with the biomedical pre-trained models. Specifically, MoE-Transformer achieves an accuracy of 0.87, which is on par with CamemBERT-bio and AliBERT and only slightly lower than DrBERT. Its precision and F1 score are both 0.87, matching the top-performing models. Although its recall is slightly lower than DrBERT, it still achieves a comparable score of 0.85. 

\begin{figure}[t]
	\centering
	\vspace{2pt}
	\includegraphics[scale=0.875]{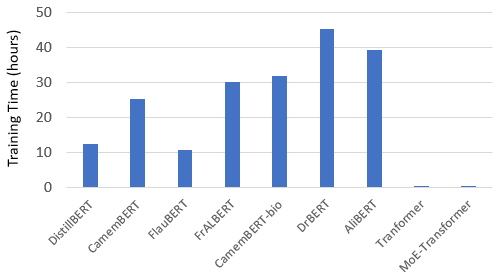}
	\vspace{-2pt}
	\caption{Training times comparison for fine-tuning the pre-trained biomedical BERT-based models vs training MoE-Transformer model.}
	\vspace{-8pt}
	\label{fig:training_times}
\end{figure}

Fig. \ref{fig:training_times} highlights the significant computational efficiency of the MoE-Transformer. The training time for the MoE-Transformer is a mere 0.17 hours, which is drastically lower than any of the pre-trained biomedical models. In contrast, DrBERT, which achieves the highest recall, requires 45.2 hours of training time (266 times longer), while CamemBERT-bio and AliBERT need 31.7 (186 times longer) and 39.3 hours (231 times longer), respectively. This apparent difference underscores the efficiency of the MoE-Transformer in terms of computational resources. It makes the MoE-Transformer highly suitable for environments with limited computational resources like hospitals. In summary, the MoE-Transformer offers a compelling balance of performance and efficiency. It achieves results on par with or better than several pre-trained and biomedical-specific models while requiring a fraction of the training time. This makes it an excellent choice for clinical applications where computational resources are constrained.

\begin{table*}[t]
\footnotesize
\centering
\caption{A comparison performance of different classifiers}
\label{tab:model_comparisions}
\begin{tabular}{|l|c|c|c|c|c|c|}
\hline
\textbf{Models} & \textbf{Accuracy} & \textbf{Precision} & \textbf{Recall} & \textbf{F1} & \textbf{Training Time (hours)} & \textbf{Inference Time (s)} \\ \hline
DistillBERT & 0.80 & 0.79 & 0.78 & 0.78 & 12.4 & 62 \\ \hline
CamemBERT & 0.83 & 0.82 & 0.83 & 0.82 & 25.1 & 132 \\ \hline
FlauBERT & 0.84 & 0.84 & 0.84 & 0.84 & 10.6 & 39 \\ \hline
FrALBERT & 0.83 & 0.82 & 0.81 & 0.81 & 30.2 & 131 \\ \hline
CamemBERT-bio & \textbf{0.87} & 0.86 & 0.88 & \textbf{0.87} & 31.7 & 142 \\ \hline
DrBERT & \textbf{0.87} & 0.84 & \textbf{0.90} & \textbf{0.87} & 45.2 & 133 \\ \hline
AliBERT & 0.86 & \textbf{0.87} & 0.84 & 0.86 & 39.3 & 128 \\ \hline
Transformer & 0.85 & 0.85 & 0.83 & 0.84 & 0.11 & 3 \\ \hline
MoETransformer & \textbf{0.87} & \textbf{0.87} & 0.85 & 0.86 & 0.17 & 4 \\ \hline
\end{tabular}
\end{table*}

Table \ref{tab:confusion_comparisions} compares the confusion matrices obtained from 9 models. Each confusion matrix presents the number of true positives (TP), false positives (FP), false negatives (FN), and true negatives (TN) for binary classification tasks. The DrBERT model achieved the highest number of correct classifications with TP+TN = 473 and the lowest number of misclassifications with FN+FP = 71. Meanwhile, the MoE-Transformer model obtained the highest number of TN (253) and the smallest number of FP (34), making it the second-best model overall. The MoE-Transformer achieved a high number of correct classifications with TP+TN = 472 and a low number of misclassifications with FN+FP = 72. This suggests that the simpler MoE-Transformer model, regarding the number of parameters, may perform comparably or even better than larger pre-trained models for a limited clinical narrative dataset.

\begin{table*}[]
\footnotesize
\centering
\caption{Confusion matrix comparison for all classifiers}
\label{tab:confusion_comparisions}
\begin{tabular}{|l|c|c|c|c|}
\hline
\textbf{Models} & \textbf{TN} $\uparrow$ & \textbf{TP} $\uparrow$ & \textbf{FP} $\downarrow$ & \textbf{FN} $\downarrow$\\ \hline
DistillBERT & 233 & 201 & 54 & 56 \\ \hline
CamemBERT & 239 & 214 & 48 & 43 \\ \hline
FlauBERT & 246 & 215 & 41 & 42 \\ \hline
FrALBERT & 241 & 209 & 46 & 48 \\ \hline
CamemBERT-bio & 228 & 243 & 40 & 33 \\ \hline
DrBERT & 238 & 235 & 46 & 25 \\ \hline
AliBERT & 235 & 231 & 44 & 44 \\ \hline
Transformer & 250 & 213 & 37 & 44 \\ \hline
MoE-Transformer & \textbf{253} & 219 & \textbf{34} & 38 \\ \hline
\end{tabular}
\end{table*}

\begin{figure}[t]
	\centering
	\vspace{2pt}
	\includegraphics[scale=0.65]{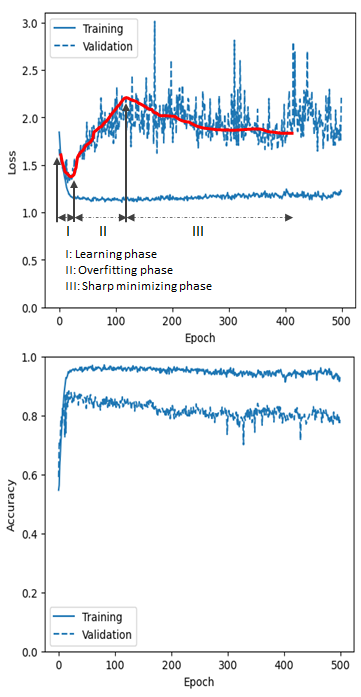}
	\vspace{-2pt}
	\caption{Generalization gap and sharp minima during training MoE-Transformer without early stopping.}
	\vspace{-8pt}
	\label{fig:sharp_minima}
\end{figure}

\begin{figure*}[t]
	\centering
	\vspace{2pt}
    \includegraphics[scale=0.525]{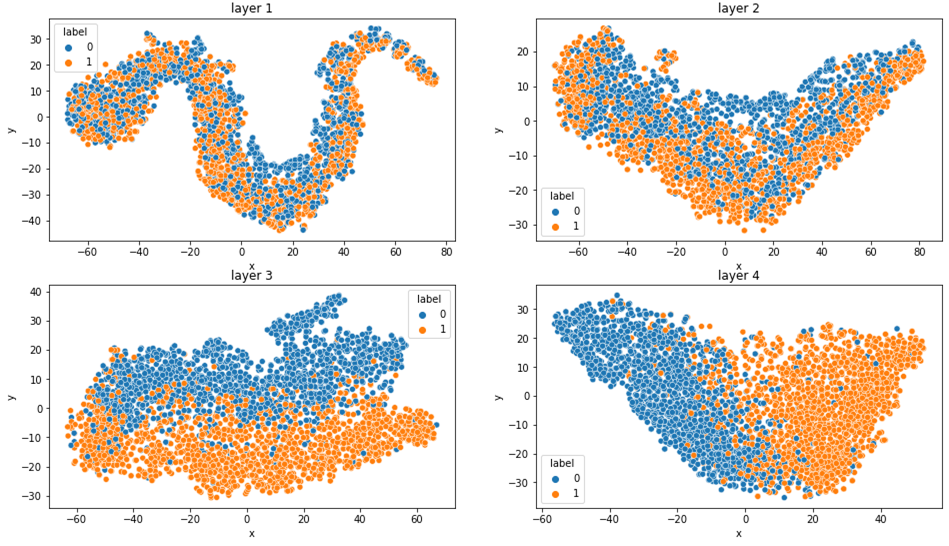}
    \includegraphics[scale=0.525]{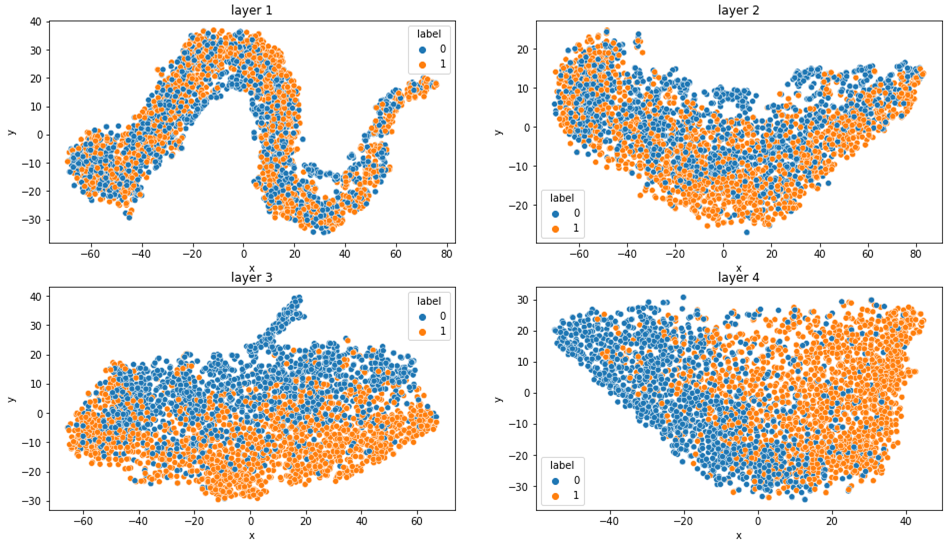}
	\vspace{-2pt}
	\caption{Hidden embedding visualization during training (top 4 figures) and validation (bottom 4 figures) for the MoE-Transformer at the 30th epoch.}
	\vspace{-8pt}
	\label{fig:hiden_train_20}
\end{figure*}

Although the MoE-Transformer outperforms several other models and the conventional Transformer model, its performance falls short when compared to two of our previous studies \cite{le2021detecting, le2023adaptation} that extensively analyzed a conceptual framework for detecting a patient's health condition from contextual input to output. On the same dataset, the proposed framework in those studies utilized a combination of TF-IDF (term frequency-inverse document frequency) and MLP-NN (multilayer perceptron neural network), achieving an overall classification performance of 89\% accuracy, 88\% recall, and 89\% precision. Moreover, sparsity reduction significantly affected classifier performance in downstream tasks, and a generative AE (autoencoder) learning algorithm effectively leveraged sparsity reduction to help the MLP-NN classifier achieve 92\% accuracy, 91\% recall, 91\% precision, and 91\% F1-score. These findings suggest that the simpler frameworks are effective for this specific context and highlight the limitations of the MoE-Transformer model.

While the MoE-Transformer model has demonstrated promising results in clinical text classification, there is still room for further improvement in its performance. One possible area of investigation is the training methodology, as suggested by previous research \cite{hoffer2017train, nakkiran2021deep}. Specifically, the model was trained for 500 epochs without early stopping, which resulted in three distinctive phases in the learning curves of training and validation losses in Fig. \ref{fig:sharp_minima}. Initially, the model underwent the learning phase, where the loss gradually decreased and reached its minimum at epoch 30. Subsequently, the model entered the second phase, where overfitting occurred, and the loss increased sharply, reaching its maximum at epoch 120. Interestingly, the model experienced a double descent, and the loss started decreasing again in the third phase and remained flat until nearly the end of the 400 epochs. During this phase, the classifier was confined to a sharp minimum and failed to improve further. Regarding accuracy, after achieving the optimal value, both learning curves from training and validation remained flat, which is expected. These are typical phenomena in deep learning models trained on small datasets, as the model tends to overfit the data and struggles with generalization. The classifier could not bridge the generalization gap caused by the sharp minima effect due to insufficient data explained in \cite{KeskarMNST17}.

Furthermore, we propose a novel perspective on this behavior and find a better illustration, viewing them through hidden embedding visualization for each layer during training and validation to explain their behavior. To illustrate this perspective, we present detailed visualizations of the MoE-Transformer embedding for each layer (from 1 to 4) in Figure \ref{fig:hiden_train_20}. We utilize t-SNE, a nonlinear dimensionality reduction technique well-suited for embedding high-dimensional data into lower-dimensional data (2 dimensions in our case). By analyzing the hidden embedding from the model, we successfully observe the difference between the training and validation processes. The four top figures illustrate that after the 30th epoch, the model successfully separates the two classes (1: positive, 0: negative) in each hidden layer. Remarkably, the last hidden layer (4th layer) achieves perfect classification accuracy of 98\% on the training set. However, this level of performance does not carry over to the validation set at the same epoch. The four bottom figures demonstrate that the two classes overlap, and the model cannot learn a clear boundary between them, resulting in only 87\% validation accuracy. Therefore, we observe a generalization gap between the training and validation for a large model with small data.

\section{Misclassification Interpretability}
\label{sec:misclass}

Interpretability of misclassifications is essential to model evaluation, particularly in critical applications such as medical diagnosis. In this study, we analyze the misclassification cases of the MoE-Transformer model by visualizing the results from the misclassification. There are  72 cases of misclassification from the results of the MoE-Transformer. Our focus has been primarily on the false negatives, where the true label indicates the presence of cardiac failure (True label is 1); however, our classifiers predict the opposite. We have referred to the labeled data to better understand the reasons behind these misclassifications. The clinician analyzes and confirms which information was inferred to label the data. 

Technically, Integrated Gradients (IG) \cite{sundararajan2017axiomatic} are a powerful interpretability technique for explaining the predictions of deep learning models, including the Transformers model used in clinical text classification. IG provides a way to attribute importance to the input features of a model, allowing clinicians and researchers to gain insight into how the model is making its predictions. Then, we compared this information with the information from the classifier based on the IG methods. This helped us identify misclassification sources and improve our classifiers' accuracy in detecting cardiac failure.

\begin{figure*}[!ht]
	\centering
	\vspace{2pt}
	\includegraphics[scale=0.525]{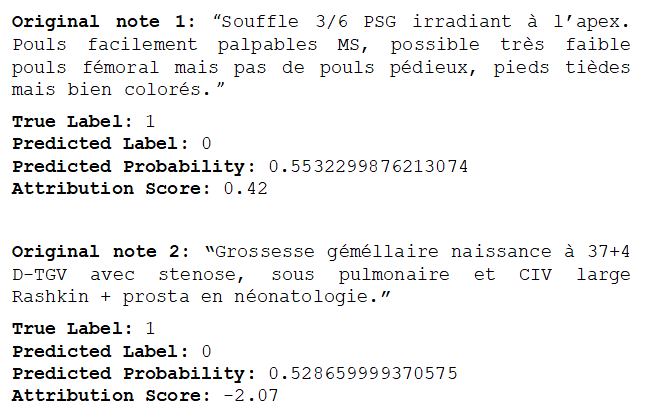}
	\vspace{-2pt}
	\caption{The highlighted misclassification cases from the MoE-Transformer model.}
	\vspace{-8pt}
	\label{fig:Misclassification}
\end{figure*}

The results in Fig. \ref{fig:Misclassification} demonstrate the Transformer model's ability to calculate attribution scores to predict output based on input features. The sign of the attribution score indicates the direction of the feature's influence on the output: a positive score means that the feature positively influences the output, while a negative score indicates a negative influence. However, the model did not perform well on the task at hand. The correct labeling of the data requires clinical expertise and professional knowledge. For example, in the first original note, the absence of data on cardiac failure was compensated for by the presence of other clinical signs such as `Souffle 3/6,' `très faible pouls fémoral mais pas de pouls pédieux (very weak femoral pulse but no pedal pulse),' and `Pieds tièdes (warm feet).' Similarly, in the second note, no data on cardiac failure was present, but `sténose sous pulmonaire et CIV large (subpulmonary stenosis and wide CIV)' suggested its presence. These examples highlight the significant gap in the Transformer model's contextual learning and understanding of real clinical datasets. There are two possible reasons for this limitation. First, while Transformer models have shown promising performance in new tasks, it remains unclear if they can generalize across the differences in settings within the clinical domain \cite{bear2021clinically}. Second, the tasks in the clinical domain often have a low signal-to-noise ratio, where the presence of a few essential keywords may suffice to determine a specific label. In contrast, Transformer's training process involves learning intricate and nuanced relations between all words in the pretraining corpus, which may not be relevant for the classification task and may shift attention away from the critical keywords \cite{gao2021limitations}. 

\section{Limitations}
While large language models (LLMs) have undoubtedly advanced numerous NLP tasks \cite{khan2025comprehensive}, their deployment in specialized clinical environments remains impractical. In our setting, stringent data privacy regulations prohibit transferring sensitive patient information to external servers or APIs, precluding fine-tuning and secure deployment of large off-the-shelf LLMs. Additionally, these models demand extensive computational resources—typically high-memory GPUs or multi-GPU clusters—far exceeding the capacity of our NVIDIA Quadro P620 (2 GB GPU memory). Furthermore, because LLMs are generally trained on broad-domain corpora, they often struggle to generalize to pediatric clinical narratives without significant retraining, amplifying data, and computational requirements. Given that our dataset comprises only 5,444 short clinical narratives, any attempt to fine-tune such large models would likely lead to overfitting and suboptimal performance. For example, a recent study \cite{le2024impact} confirms that employing Low-Rank Adaptation during fine-tuning has proven ineffective in overcoming these limitations.  Hence, a more focused and computationally efficient model, such as our MoE-Transformer, is a more viable alternative under these constraints.

Similarly, although state-of-the-art LLMs like Mistral \cite{jiang2023mistral} and LLaMA 3.21b \cite{dubey2024llama} have demonstrated impressive capabilities in general NLP tasks, several fundamental limitations hinder their integration into our CDSS at CHUSJ. First, even the smallest LLMs typically contain no fewer than 5 billion parameters \cite{lu2024small}, making them more than 40 times larger than our domain-specific models, specifically trained on French biomedical clinical texts. This vast difference in scale implies that, even in zero-shot settings, general-purpose LLMs are prone to hallucinations and unreliable predictions when applied to pediatric clinical narratives \cite{omiye2024large}. Second, while these models are trained on massive datasets ranging from 825 billion to 6.3 trillion tokens, our dataset comprises only 580,000 tokens, further questioning their effectiveness without extensive domain-specific fine-tuning. Third, the computational demands of even the smallest LLMs (e.g., requiring a minimum of 1024-core NVIDIA Ampere architecture GPU with 32 tensor cores, 16G DRAM) are incompatible with our hardware capabilities. Finally, strict hospital privacy regulations preclude cloud-based inference, ruling out the possibility of relying on externally hosted models or resource-intensive fine-tuning methods. Despite their success in broader NLP applications, LLMs are not a feasible solution for our specialized clinical setting, thereby underscoring the advantages of our MoE-Transformer for local, real-time clinical text classification.

\section{Conclusion}
\label{sec:conclusion}

We compared the performance of 9 classifiers on a binary classification task. The results indicated that MoE can improve the performance of Transformer models over pre-trained BERT-based models. The MoE-Transformer model performed comparable to biomedical pre-trained models but with 100 times less computation resources. These results confirm that the MoE-Transformer is particularly valuable for classifying small French clinical narratives within the privacy and constraints of hospital-based computational resources.

The study used attribution scores to demonstrate the MoE-Transformer model's ability to predict output based on input features. However, the model did not perform well on the clinical dataset due to its inability to contextualize and understand real-world data. The clinical tasks have a low signal-to-noise ratio, and the MoE-Transformer's training process may shift attention away from critical keywords. Additionally, it remains unclear whether Transformer models can generalize across different settings in the clinical domain. Overall, the results suggest the need for further research to improve the MoE-Transformer model's performance in clinical settings.

These findings suggest that carefully training the Transformer-based models from scratch can significantly improve the performance of clinical narrative classification tasks. The CDSS at CHUSJ is especially currently under development. By combining this NLP algorithm to detect the absence of heart failure with the two other algorithms already developed on hypoxemia detection \cite{sauthier2021estimated} and chest, X-ray analysis \cite{zaglam2014computer, yahyatabar2020dense}, the next step of our study is to implement the resulting CDSS (integration of the three algorithms) within the cyberinfrastructure of the pediatric intensive care unit (PICU) at Sainte-Justine Hospital to diagnose ARDS early. We will verify the CDSS's ability to detect ARDS prospectively once the integration with the PICU e-medical infrastructure is completed.

Future work will further explore enhancements to the evaluation methodology to gain deeper insights into the model's performance and fairness in clinical text classification. Specifically, plans include incorporating class-wise performance metrics \cite{benz2021robustness} to reveal detailed strengths and weaknesses across different clinical categories, thereby identifying potential biases and underperforming segments. Complementing this, a qualitative evaluation framework employing text perturbation experiments—such as synonym replacements, misspellings, and grammatical variations—will be developed to assess the robustness and generalization capabilities of the MoE-Transformer compared to large language model baselines. Additionally, drawing on insights from recent work on extracting social determinants of health from clinical texts \cite{raza2023discovering}, comparative analyses will be conducted to benchmark this approach against state-of-the-art methodologies. Recognizing the critical importance of fairness, future studies will also systematically examine potential biases in data curation, compilation, and model development by leveraging established frameworks like the AI Risk Repository, as highlighted in \cite{slattery2024ai}. These explorations are expected to validate the approach's practicality and robustness while contributing to developing fair, responsible, and practical AI applications in healthcare.

\section*{Acknowledgment}

This work was supported by a scholarship from the Fonds de recherche du Quebec-Nature et technologies (FRQNT) to Thanh-Dung Le, and grants from the Natural Sciences and Engineering Research Council (NSERC), the Institut de valorization des donnees (IVADO), and the Fonds de la recherche en sante du Quebec (FRQS). Data and reproducible codes are available upon request from Prof. Philippe Jouvet, M.D., PhD (Email: philippe.jouvet.med@ssss.gouv.qc.ca).

\bibliographystyle{IEEEtran}
\flushend
\bibliography{IEEEabrv,Bibliography}
\end{document}